\documentclass[11pt,a4paper]{article}
\usepackage[hyperref]{naaclhlt2018}
\usepackage{times}
\usepackage{latexsym}
\usepackage{graphicx}
\usepackage{url}
\usepackage{booktabs}
\usepackage{natbib}
\usepackage{amssymb}
\usepackage{pifont}
\usepackage[textwidth=1.5cm]{todonotes} 

\aclfinalcopy 
\def\aclpaperid{***} 

\newcommand{\cross}{\textcolor{red}{\ding{54}}}
\newcommand{\cmark}{\textcolor{green}{\ding{52}}}

\title{UG18 at SemEval-2018 Task 1: Generating Additional Training Data for Predicting Emotion Intensity in Spanish}

\author{Marloes Kuijper \\
  CLCG \\
  University of Groningen \\
  {\tt marloes.madelon} \\
  {\tt @gmail.com} \\\And
  Mike van Lenthe \\
  CLCG \\
  University of Groningen \\
  {\tt mikevanlenthe} \\
  {\tt @gmail.com}
  \\\And
    Rik van Noord \\
  CLCG \\
  University of Groningen \\
  {\tt r.i.k.van.noord@rug.nl}}

\date{}

\begin{document}
\maketitle
\begin{abstract}
The present study describes our submission to SemEval 2018 Task 1: Affect in Tweets. Our Spanish-only approach aimed to demonstrate that it is beneficial to automatically generate additional training data by (i) translating training data from other languages and (ii) applying a semi-supervised learning method. We find strong support for both approaches, with those models outperforming our regular models in all subtasks. However, creating a stepwise ensemble of different models as opposed to simply averaging did not result in an increase in performance. We placed second (EI-Reg), second (EI-Oc), fourth (V-Reg) and fifth (V-Oc) in the four Spanish subtasks we participated in.
\end{abstract}

\section{Introduction}

\begin{table*}[htb]
\centering
\resizebox{\columnwidth * 2}{!}{%
\begin{tabular}{l|ccccccccccccc}
\toprule
\textbf{Task} & \textbf{NRC-HSL} & \textbf{S-140} & \textbf{SenStr} & \textbf{AFINN} & \textbf{EMOTICONS}  & \textbf{Bing Liu} & \textbf{MPQA} & \textbf{NRC-10-exp} & \textbf{NRC-HEAL} & \textbf{NEGATION}  \\
\toprule
\textit{EI-Reg-a}  & \cmark             & \cmark              & \cross               & \cmark               & \cmark                            & \cross             & \cmark                      & \cmark          & \cross              & \cross                \\
\textit{EI-Reg-f}    & \cross             & \cross              & \cross               & \cmark               & \cross                            & \cmark             & \cmark                      & \cross          & \cmark              & \cmark                   \\
\textit{EI-Reg-j}     & \cmark             & \cmark              & \cmark               & \cross               & \cross                            & \cmark             & \cross                      & \cross          & \cross              & \cmark                    \\
\textit{EI-Reg-s} & \cross             & \cmark              & \cross               & \cross               & \cmark                            & \cmark             & \cross                      & \cmark          & \cmark              & \cross               \\
\textit{EI-Oc-a}    & \cross             & \cmark              & \cross               & \cross               & \cross                           & \cross             & \cmark                     & \cross          & \cmark              & \cross                  \\
\textit{EI-Oc-f}    & \cmark             & \cross              & \cross               & \cross               & \cross                           & \cross             & \cross                     & \cross          & \cross              & \cmark                   \\
\textit{EI-Oc-j}      & \cross             & \cross              & \cross               & \cross               & \cross                          & \cmark             & \cmark                      & \cmark          & \cross              & \cmark                   \\
\textit{EI-Oc-s}  & \cross             & \cross              & \cmark               & \cmark               & \cmark                            & \cross             & \cross                      & \cmark          & \cmark              & \cross                    \\
\textit{V-Reg}          & \cross             & \cross              & \cross               & \cross               & \cross                            & \cmark             & \cmark                      & \cmark          & \cross              & \cross                   \\
\textit{V-Oc}           & \cross             & \cross              & \cross               & \cross               & \cross                           & \cross             & \cmark                      & \cross          & \cross              & \cross                    \\

\bottomrule
\end{tabular}%
}
\caption{\label{lexicons}Lexicons included in our final ensemble. NRC-10 and SentiWordNet are left out of the table because they never improved the score for a task.}
\end{table*}


Understanding the emotions expressed in a text or message is of high relevance nowadays. Companies are interested in this to get an understanding of the sentiment of their current customers regarding their products and the sentiment of their potential customers to attract new ones. Moreover, changes in a product or a company may also affect the sentiment of a customer. However, the intensity of an emotion is crucial in determining the urgency and importance of that sentiment. If someone is only slightly happy about a product, is a customer willing to buy it again? Conversely, if someone is very angry about customer service, his or her complaint might be given priority over somewhat milder complaints.


\citet{SemEval2018Task1} present four tasks\footnote{We did not participate in subtask 5 (E-c).} in which systems have to automatically determine the intensity of emotions (EI) or the intensity of the sentiment (Valence) of tweets in the languages English, Arabic, and Spanish. The goal is to either predict a continuous regression (reg) value or to do ordinal classification (oc) based on a number of predefined categories. The EI tasks have separate training sets for four different emotions: anger, fear, joy and sadness. Due to the large number of subtasks and the fact that this language does not have many resources readily available, we only focus on the Spanish subtasks. Our work makes the following contributions:


\begin{itemize}
	
	\item{We show that automatically translating English lexicons and English training data boosts performance;}
    \item{We show that employing semi-supervised learning is beneficial;}
    \item{We show that the stepwise creation of an ensemble model is not necessarily better method than simply averaging predictions.}
\end{itemize}    


Our submissions ranked second (EI-Reg), second (EI-Oc), fourth (V-Reg) and fifth (V-Oc), demonstrating that the proposed method is accurate in automatically determining the intensity of emotions and sentiment of Spanish tweets. 
This paper will first focus on the datasets, the data generation procedure, and the techniques and tools  used. Then we present the results in detail, after which we perform a small error analysis on the largest mistakes our model made. We conclude with some possible ideas for future work.

\section{Method}

\subsection{Data}
\label{sec:data}
For each task, the training data that was made available by the organizers is used, which is a selection of tweets with for each tweet a label describing the intensity of the emotion or sentiment \citep{LREC18-TweetEmo}. Links and usernames were replaced by the general tokens \texttt{URL} and \texttt{@username}, after which the tweets were tokenized by using TweetTokenizer. All text was lowercased. In a post-processing step, it was ensured that each emoji is tokenized as a single token.

\subsection{Word Embeddings}

To be able to train word embeddings, Spanish tweets were scraped between November 8, 2017 and January 12, 2018. We chose to create our own embeddings instead of using pre-trained embeddings, because this way the embeddings would resemble the provided data set: both are based on Twitter data. Added to this set was the Affect in Tweets Distant Supervision Corpus (DISC) made available by the organizers \citep{SemEval2018Task1} and a set of 4.1 million tweets from 2015, obtained from \citet{toral2015dublin}. After removing duplicate tweets and tweets with fewer than ten tokens, this resulted in a set of 58.7 million tweets, containing 1.1 billion tokens. The tweets were preprocessed using the method described in Section~\ref{sec:data}. The word embeddings were created using word2vec in the gensim library \cite{rehurek_lrec}, using CBOW, a window size of 40 and a minimum count of 5.\footnote{Embeddings available at \url{www.let.rug.nl/rikvannoord/embeddings/spanish/}} 
The feature vectors for each tweet were then created by using the AffectiveTweets WEKA package \cite{MohammadB17}.

\subsection{Translating Lexicons}

Most lexical resources for sentiment analysis are in English. To still be able to benefit from these sources, the lexicons in the AffectiveTweets package were translated to Spanish, using the machine translation platform Apertium \cite{forcada2011apertium}. 


All lexicons from the AffectiveTweets package were translated, except for SentiStrength. Instead of translating this lexicon, the English version was replaced by the Spanish variant made available by \citet{bravo2013combining}.




For each subtask, the optimal combination of lexicons was determined. This was done by first calculating the benefits of adding each lexicon individually, after which only beneficial lexicons were added until the score did not increase anymore (e.g. after adding the best four lexicons the fifth one did not help anymore, so only four were added). The tests were performed using a default SVM model, with the set of word embeddings described in the previous section. Each subtask thus uses a different set of lexicons (see Table \ref{lexicons} for an overview of the lexicons used in our final ensemble). For each subtask, this resulted in a (modest) increase on the development set, between 0.01 and 0.05.




\subsection{Translating Data}

The training set provided by \citet{SemEval2018Task1} is not very large, so it was interesting to find a way to augment the training set. A possible method is to simply translate the datasets into other languages, leaving the labels intact. Since the present study focuses on Spanish tweets, all tweets from the English datasets were translated into Spanish. This new set of ``Spanish'' data was then added to our original training set. Again, the machine translation platform Apertium \cite{forcada2011apertium} was used for the translation of the datasets. 

\subsection{Algorithms Used}
Three types of models were used in our system, a feed-forward neural network, an LSTM network and an SVM regressor. The neural nets were inspired by the work of Prayas \cite{goel2017prayas} in the previous shared task. Different regression algorithms (e.g. AdaBoost, XGBoost) were also tried due to the success of SeerNet \cite{duppada-hiray:2017:WASSA2017}, but our study was not able to reproduce their results for Spanish.

\begin{table*}[!ht]
\centering

\begin{tabular}{l|c|cc|cccc}
\toprule
\textbf{}     & \textbf{SVM}     & \multicolumn{2}{c}{\textbf{Feed-forward}} & \multicolumn{4}{|c}{\textbf{LSTM}}                                    \\
\midrule
\textbf{Task} & \textbf{Epsilon} & \textbf{Layers}     & \textbf{Nodes}      & \textbf{Layers} & \textbf{Nodes} & \textbf{Dropout} & \textbf{Dense} \\ \midrule
\textit{EI-Reg-a}      & 0.01             & 2                   & (600, 200)          & 2               & 400            & 0.001            &  \cross              \\
\textit{EI-Reg-f}     & 0.04             & 2                   & (700, 200)          & 2               & 400            & 0.01             &  \cmark              \\
\textit{EI-Reg-j}      & 0.05             & 2                   & (500, 500)          & 2               & 200            & 0.1              &  \cross              \\
\textit{EI-Reg-s}      & 0.06             & 2                   & (400, 300)          & 2               & 600            & 0.001            & \cross               \\
\textit{EI-Oc-a}       & 0.005            & 2                   & (600, 200)          & 2               & 200            & 0.001            & \cross                \\
\textit{EI-Oc-f}       & 0.06             & 2                   & (700, 300)          & 2               & 200            & 0.001            &  \cross              \\
\textit{EI-Oc-j}       & 0.04             & 2                   & (800, 200)          & 3               & 400            & 0.001            & \cmark               \\
\textit{EI-Oc-s}       & 0.005            & 2                   & (500, 200)          & 3               & 800            & 0.01             & \cmark               \\
\textit{V-Reg}         & 0.07             & 3                   & (400, 400, 400)     & 2               & 200            & 0.001            &  \cmark              \\
\textit{V-Oc}         & 0.09             & 3                   & (400, 400, 100)     & 3               & 600            & 0.01             &  \cmark \\ \bottomrule              
\end{tabular}
\caption{Parameter settings for the algorithms used. For feed-forward, we show the number of nodes per layer. The \textit{Dense} column for LSTM shows whether a dense layer was added after the LSTM layers (with half the number of nodes as is shown in the \textit{Nodes} column). The feed-forward networks always use a dropout of 0.001 after the first layer.}\label{tab-params}
\end{table*}

For both the LSTM network and the feed-forward network, a parameter search was done for the number of layers, the number of nodes and dropout used. This was done for each subtask, i.e. different tasks can have a different number of layers. All models were implemented using Keras \cite{chollet2015keras}. After the best parameter settings were found, the results of 10 system runs to produce our predictions were averaged (note that this is different from averaging our different \textit{type} of models in Section \ref{sec:ensembling}). For the SVM (implemented in scikit-learn \citep{scikit-learn}), the RBF kernel was used and a parameter search was conducted for epsilon. Detailed parameter settings for each subtask are shown in Table \ref{tab-params}. Each parameter search was performed using 10-fold cross validation, as to not overfit on the development set.

\subsection{Semi-supervised Learning}

\begin{table*}[!htb]
\centering
\begin{tabular}{lccl|cc|cc}
\toprule
                 &                &                 &               & \multicolumn{2}{c|}{\textbf{Feed-forward}}  & \multicolumn{2}{c|}{\textbf{LSTM}}          \\ \midrule
                 & \textbf{Words} & \textbf{Tweets} & \textbf{Task} & \textbf{Threshold} & \textbf{Tweets added} & \textbf{Threshold} & \textbf{Tweets added} \\ \midrule
\textbf{Anger}   & 23             & 81, 798         & \textit{EI-Reg}       & 0.1                & 2,500                  & 0.05               & 2,500                  \\
                 &                &                 & \textit{EI-Oc}         & 0.1                & 1,000                  & 0.1                & 2,500                  \\
\textbf{Fear}    & 17             & 54,113          & \textit{EI-Reg}        & 0.1                & 1,500                  & 0.05               & 1,500                  \\
                 &                &                 & \textit{EI-Oc}         & 0.075              & 1,000                  & 0.1                & 2,500                  \\
\textbf{Joy}     & 29             & 51,135          & \textit{EI-Reg}       & 0.125              & 1,500                  & 0.15               & 500                   \\
                 &                &                 & \textit{EI-Oc}        & 0.05               & 500                   & 0.05               & 500                   \\
\textbf{Sadness} & 16             & 102,810         & \textit{EI-Reg}        & 0.1                & 5,000                  & 0.1                & 2,500                  \\
                 &                &                 & \textit{EI-Oc}         & 0.125              & 2,000                  & 0.05               & 2,500        \\ \bottomrule         
\end{tabular}
\caption{Statistics and parameter settings of the semi-supervised learning experiments.}\label{tab:semi}
\end{table*}

One of the aims of this study was to see if using semi-supervised learning is beneficial for emotion intensity tasks. For this purpose, the DISC \citep{SemEval2018Task1} corpus was used. This corpus was created by querying certain emotion-related words, which makes it very suitable as a semi-supervised corpus. However, the specific emotion the tweet belonged to was not made public. Therefore, a method was applied to automatically assign the tweets to an emotion by comparing our scraped tweets to this new data set.

First, in an attempt to obtain the query-terms, we selected the 100 words which occurred most frequently in the DISC corpus, in comparison with their frequencies in our own scraped tweets corpus. Words that were clearly not indicators of emotion were removed. The rest was annotated per emotion or removed if it was unclear to which emotion the word belonged. This allowed us to create \textit{silver} datasets per emotion, assigning tweets to an emotion if an annotated \textit{emotion-word} occurred in the tweet. 

Our semi-supervised approach is quite straightforward: first a model is trained  on the training set and then this model is used to predict the labels of the silver data. This silver data is then simply added to our training set, after which the model is retrained. However, an extra step is applied to ensure that the silver data is of reasonable quality. Instead of training a single model initially, ten different models were trained which predict the labels of the silver instances. If the highest and lowest prediction do not differ more than a certain threshold the silver instance is maintained, otherwise it is discarded. 

This results in two parameters that could be optimized: the threshold and the number of silver instances that would be added. This method can be applied to both the LSTM and feed-forward networks that were used. An overview of the characteristics of our data set with the final parameter settings is shown in Table~\ref{tab:semi}. Usually, only a small subset of data was added to our training set, meaning that most of the silver data is not used in the experiments. Note that since only the emotions were annotated, this method is only applicable to the EI tasks.\footnote{For EI-Oc, the labels were normalized between 0 and 1.}

\subsection{Ensembling}
\label{sec:ensembling}
To boost performance, the SVM, LSTM, and feed-forward models were combined into an ensemble. For both the LSTM and feed-forward approach, three different models were trained. The first model was trained on the training data (regular), the second model was trained on both the training and translated training data (translated) and the third one was trained on both the training data and the semi-supervised data (silver). Due to the nature of the SVM algorithm, semi-supervised learning does not help, so only the regular and translated model were trained in this case. This results in 8 different models per subtask. Note that for the valence tasks no silver training data was obtained, meaning that for those tasks the semi-supervised models could not be used.

Per task, the LSTM and feed-forward model's predictions were averaged over 10 prediction runs. Subsequently, the predictions of all individual models were combined into an average. Finally, models were removed from the ensemble in a stepwise manner if the removal increased the average score. This was done based on their original scores, i.e. starting out by trying to remove the worst individual model and working our way up to the best model. We only consider it an increase in score if the difference is larger than 0.002 (i.e. the difference between 0.716 and 0.718). If at some point the score does not increase and we are therefore unable to remove a model, the process is stopped and our best ensemble of models has been found. This process uses the scores on the development set of different combinations of models. Note that this means that the ensembles for different subtasks can contain different sets of models. The final model selections can be found in Table~\ref{models}. 



\begin{table*}[!ht]
\centering
\begin{tabular}{l|cccccccc}
\toprule
\textbf{Task}           & \textbf{SVM-r} & \textbf{SVM-t} & \textbf{LSTM-r} & \textbf{LSTM-t} & \textbf{LSTM-s} & \textbf{FF-r} & \textbf{FF-t} & \textbf{FF-s} \\
\toprule
\textit{EI-Reg-anger}   & \cross              & \cmark              & \cross               & \cmark               & \cmark               & \cmark             & \cmark             & \cmark             \\
\textit{EI-Reg-fear}    & \cmark              & \cmark              & \cross               & \cmark               & \cmark               & \cross             & \cmark             & \cross             \\
\textit{EI-Reg-joy}     & \cmark              & \cmark              & \cross               & \cmark               & \cmark               & \cross             & \cmark             & \cmark             \\
\textit{EI-Reg-sadness} & \cmark              & \cmark              & \cmark               & \cross               & \cross               & \cmark             & \cmark             & \cmark             \\
\textit{EI-Oc-anger}    & \cmark              & \cross              & \cross               & \cmark               & \cmark               & \cross             & \cmark             & \cmark             \\
\textit{EI-Oc-fear}     & \cross              & \cmark              & \cmark               & \cmark               & \cmark               & \cmark             & \cmark             & \cmark             \\
\textit{EI-Oc-joy}      & \cross              & \cmark              & \cross               & \cmark               & \cmark               & \cmark             & \cmark             & \cmark             \\
\textit{EI-Oc-sadness}  & \cross              & \cmark              & \cross               & \cmark               & \cmark               & \cmark             & \cmark             & \cmark             \\
\textit{V-Reg}          & \cmark              & \cross              & \cross               & \cmark               & \cross               & \cmark             & \cmark             & \cross    \\
\textit{V-Oc}           & \cross              & \cmark              & \cmark               & \cmark               & \cross               & \cmark             & \cmark             & \cross             
 
\\ \bottomrule
\end{tabular}
\caption{\label{models}Models included in our final ensemble.}
\end{table*}

\begin{table*}[!ht]
\centering
\begin{tabular}{l|cccccccc}
\toprule
\textbf{Task}           & \textbf{SVMr} & \textbf{SVMt} & \textbf{LSTMr} & \textbf{LSTMt} & \textbf{LSTMs} & \textbf{FFr} & \textbf{FFt} & \textbf{FFs} \\ 
\toprule
\textit{EI-Reg-a}   & 0.630         & 0.663         & 0.644          & 0.672          & 0.683          & 0.659        & 0.672        & \textbf{0.681}     \\  
\textit{EI-Reg-f}    & 0.683         & 0.700         & 0.666          & 0.702           & 0.682           & 0.675        & \textbf{0.704}        & 0.674  \\ 
\textit{EI-Reg-j}     & 0.702         & 0.711         & 0.683           & 0.709         & 0.699          & 0.688        & \textbf{0.720}        & 0.710   \\  
\textit{EI-Reg-s} & 0.690         & 0.696         & 0.694          & 0.67            & 0.678          & 0.694         & 0.694        & \textbf{0.704}  \\ 
\textit{EI-Oc-a}    & 0.663          & 0.645         & 0.602          & \textbf{0.673}          & 0.589          & 0.611        & 0.659        & 0.640    \\ 
\textit{EI-Oc-f}     & 0.621         & 0.579          & 0.610          & 0.603          & 0.615          & 0.596         & 0.598        & \textbf{0.629}     \\  
\textit{EI-Oc-j}      & 0.626          & \textbf{0.674}         & 0.670          & 0.657          & 0.671          & 0.616        & 0.638        & 0.628    \\  
\textit{EI-Oc-s}  & 0.579          & 0.621         & 0.590          & 0.612          & 0.610          & 0.579        & \textbf{0.633}        & 0.595    \\ 
\textit{V-Reg}          & 0.728         & 0.735          & 0.729          & 0.766          & -               & 0.751        & \textbf{0.765}        & -        
\\ 
\textit{V-Oc}           & 0.680         & 0.670         & 0.719          & 0.711          & -               & 0.724        & \textbf{0.727}        & -      \\  
\bottomrule
\end{tabular}
\caption{Scores for each individual model per subtask. Best individual score per subtask is bolded.}
\label{scores}
\end{table*}

\begin{table}[ht]
\centering
\begin{tabular}{l|cccc}
\toprule
\textbf{Task} & \textbf{\begin{tabular}[c]{@{}c@{}}Avg \\ Dev\end{tabular}} & \textbf{\begin{tabular}[c]{@{}c@{}}Ens \\ Dev\end{tabular}} & \textbf{\begin{tabular}[c]{@{}c@{}}Avg \\ Test\end{tabular}} & \textbf{\begin{tabular}[c]{@{}c@{}}Ens \\ Test\end{tabular}} \\ \midrule
\textit{EI-Reg-a}      & 0.684            & 0.692            & 0.589             & 0.595             \\
\textit{EI-Reg-f}      & 0.709            & 0.718            & 0.687             & 0.689             \\
\textit{EI-Reg-j}      & 0.721            & 0.727            & 0.712             & 0.712             \\
\textit{EI-Reg-s}     & 0.711            & 0.716            & 0.710             & 0.712             \\
\textit{EI-Oc-a}       & 0.658            & 0.678            & 0.500             & 0.499             \\
\textit{EI-Oc-f}       & 0.643            & 0.666            & 0.592             & 0.606             \\
\textit{EI-Oc-j}      & 0.669            & 0.695            & 0.668             & 0.665             \\
\textit{EI-Oc-s}       & 0.612            & 0.645            & 0.612             & 0.625             \\
\textit{V-Reg}        & 0.728            & 0.744            & 0.686             & 0.682             \\
\textit{V-Oc}          & 0.767            & 0.772            & 0.706             & 0.707             \\
\bottomrule
\end{tabular}

\caption{Results on the dev and test set for averaging and stepwise ensembling the individual models. The last column shows our official results.}
\label{tab:official}
\end{table}

\section{Results and Discussion}
Table \ref{scores} shows the results on the development set of all individuals models, distinguishing the three types of training: regular (r), translated (t) and semi-supervised (s). In Tables \ref{models} and \ref{scores}, the letter behind each model (e.g. SVM-r, LSTM-r) corresponds to the type of training used. Comparing the regular and translated columns for the three algorithms, it shows that in 22 out of 30 cases, using translated instances as extra training data resulted in an improvement. For the semi-supervised learning approach, an improvement is found in 15 out of 16 cases. Moreover, our best individual model for each subtask (bolded scores in Table \ref{scores}) is always either a translated or semi-supervised model. Table~\ref{scores} also shows that, in general, our feed-forward network obtained the best results, having the highest F-score for 8 out of 10 subtasks.

However, Table~\ref{tab:official} shows that these scores can still be improved by averaging or ensembling the individual models. On the dev set, averaging our 8 individual models results in a better score for 8 out of 10 subtasks, while creating an ensemble beats all of the individual models as well as the average for each subtask. On the test set, however, only a small increase in score (if any) is found for stepwise ensembling, compared to averaging. Even though the results do not get worse, we cannot conclude that stepwise ensembling is a better method than simply averaging. 

Our official scores (column \textit{Ens Test} in Table~\ref{tab:official}) have placed us second (EI-Reg, EI-Oc), fourth (V-Reg) and fifth (V-Oc) on the SemEval AIT-2018 leaderboard. However, it is evident that the results obtained on the test set are not always in line with those achieved on the development set. Especially on the anger subtask for both EI-Reg and EI-Oc, the scores are considerably lower on the test set in comparison with the results on the development set. Therefore, a small error analysis was performed on the instances where our final model made the largest errors.

\begin{table*}[ht]
\centering
\resizebox{\columnwidth * 2}{!}{%
\begin{tabular}{lccl}
\toprule
\textbf{Example sentence} & \textbf{Pred.} & \textbf{Gold} & \textbf{Possible problem} \\
\midrule
QUIERES PELEA FÍSICA?                                         & 0.25                           & 0.80                      & Capitalization                          \\
DO YOU WANT A PHYSICAL FIGHT? & & & \\ \midrule
Ojalá una precuela de Imperator Furiosa.                      & 0.64                           & 0.24                      & Named entity not recognized    \\
I wish a prequel to Imperator Furiosa. & & & \\ \midrule
Odio estar tan enojada y que me de risa                     & 0.79                           & 0.46                      & Reduced angriness \\ 
I hate being so angry and that that makes me laugh & & & \\ \midrule
Yo la mejor y que te contesten así nomas me infla la vena & 0.45                           & 0.90                      & Figurative speech                       \\
I am the best and that they answer you like that, it just inflates my vein & & & \\ \bottomrule            
\end{tabular}}
\caption{Error analysis for the EI-Reg-anger subtask, with English translations.}\label{error-analysis}
\end{table*}

\subsection{Error Analysis}
Due to some large differences between our results on the dev and test set of this task, we performed a small error analysis in order to see what caused these differences.  For \textit{EI-Reg-anger}, the gold labels were compared to our own predictions, and we manually checked 50 instances for which our system made the largest errors. 

Some examples that were indicative of the shortcomings of our system are shown in Table \ref{error-analysis}. First of all, our system did not take into account capitalization. The implications of this are shown in the first sentence, where capitalization intensifies the emotion used in the sentence. In the second sentence, the name \textit{Imperator Furiosa} is not understood. Since our texts were lowercased, our system was unable to capture the named entity and thought the sentence was about an angry emperor instead. In the third sentence, our system fails to capture that when you are so angry that it makes you laugh, it results in a reduced intensity of the angriness. Finally, in the fourth sentence, it is the figurative language \textit{me infla la vena} (it inflates my vein) that the system is not able to understand. 

The first two error-categories might be solved by including smart features regarding capitalization and named entity recognition. However, the last two categories are problems of natural language understanding and will be very difficult to fix.

\section{Conclusion}

To conclude, the present study described our submission for the Semeval 2018 Shared Task on Affect in Tweets. We participated in four Spanish subtasks and our submissions ranked second, second, fourth and fifth place.
Our study aimed to investigate whether the automatic generation of additional training data through translation and semi-supervised learning, as well as the creation of stepwise ensembles, increase the performance of our Spanish-language models. Strong support was found for the translation and semi-supervised learning approaches; our best models for all subtasks use either one of these approaches. 
These results suggest that both of these additional data resources are beneficial when determining emotion intensity (for Spanish). 
However, the creation of a stepwise ensemble from the best models did not result in better performance compared to simply averaging the models. In addition, some signs of overfitting on the dev set were found. In future work, we would like to apply the methods (translation and semi-supervised learning) used on Spanish on other low-resource languages and potentially also on other tasks.

\bibliography{main}
\bibliographystyle{acl_natbib}

\end{document}